\pdfoutput=1

\documentclass[11pt]{article}

\usepackage[final]{acl}

\usepackage{times}
\usepackage{latexsym}
\usepackage{float}
\usepackage{dblfloatfix}
\usepackage{comment}

\usepackage{graphicx}
\usepackage{xcolor}

\usepackage{enumitem}

\usepackage[T1]{fontenc}

\usepackage[utf8]{inputenc}

\usepackage{microtype}

\usepackage{inconsolata}
\title{A LLM-Based Ranking Method for the Evaluation \\of Automatic Counter-Narrative Generation}

\author{Irune Zubiaga \and Aitor Soroa \and Rodrigo Agerri \\
         HiTZ Center - Ixa, University of the Basque Country UPV/EHU\\ \{irune.zubiaga, a.soroa, rodrigo.agerri\}@ehu.eus}

\begin{document}
\maketitle
\begin{abstract}

  This paper proposes a novel approach to evaluate Counter Narrative (CN) generation using a Large Language Model (LLM) as an evaluator. We show that traditional automatic metrics correlate poorly with human judgements and fail to capture the nuanced relationship between generated CNs and human perception. To alleviate this, we introduce a model ranking pipeline based on pairwise comparisons of generated CNs from different models, organized in a tournament-style format. The proposed evaluation method achieves a high correlation with human preference, with a $\rho$ score of $0.88$. As an additional contribution, we leverage LLMs as zero-shot CN generators and provide a comparative analysis of chat, instruct, and base models, exploring their respective strengths and limitations. Through meticulous evaluation, including fine-tuning experiments, we elucidate the differences in performance and responsiveness to domain-specific data. We conclude that chat-aligned models in zero-shot are the best option for carrying out the task, provided they do not refuse to generate an answer due to security concerns.

\textit{\textbf{Warning:} Please be advised that this research paper contains instances of hate speech that may be distressing or offensive to readers. These expressions are included for analysis and critique purposes only, and they do not reflect the beliefs or endorsements of the authors or the institution.}
\end{abstract}

\section{Introduction}

The proliferation of misinformation and the dissemination of harmful narratives has stressed the urgent need for effective strategies to combat Hate Speech (HS). This necessity has drawn significant attention to the field of automatic CN generation, where considerable research has focused on the use of LLMs to fulfill this task \citep{chung-etal-2021-towards,tekiroglu-etal-2022-using}. However, difficulties in automatically assessing the quality of the generated CNs remain. As is common in text generation tasks, while manual evaluation is expensive, time-consuming, and subjective, existing automatic methods often fail to provide comprehensive insights or capture the nuanced relationship between generated text and human perception, overlooking crucial aspects of effectiveness and relevance~\citep{nimah2023nlg}. 
Finally, the problem of CN evaluation is exacerbated by the lack of a 'universal truth' and the significant variations among possible references, as shown in Table~\ref{tab:CONAN-ex}.


In this paper we address the limitations of traditional evaluation metrics for CN generation by proposing a novel automatic evaluation approach. This method is motivated by the need to improve upon existing metrics like BLEU, ROUGE and BERTScore~\citep{papineni-etal-2002-bleu,lin-2004-rouge,zhang2020bertscore} which do not consider the specific HS to which the CN is responding to, an essential aspect for accurately assessing the quality of CNs. Our approach evaluates generated CNs pairwise in a tournament-style format, with outcomes determined without human intervention using JudgeLM~\citep{zhu2023judgelm}, a model explicitly trained to assess the quality of text. Thus, JudgeLM enables pairwise comparisons of CNs, addressing the subjectivity of the task by breaking it down into simpler binary classification problems. To ensure the generalizability of our approach for the task, we test it on two distinct corpora: CONAN~\citep{chung2019conan} and CONAN-MT~\citep{fanton2021human}. Using various models to generate texts of different quality means that we can evaluate model performance across a spectrum of CN quality, ensuring that even subtle distinctions between good texts can be captured. This approach ultimately aims for a higher correlation to human preference than traditional metrics and ranks models based on their ability to generate high-quality CNs.


As an additional contribution, we evaluate the inherent ability of LLMs as zero shot (ZS) CN generators. By leveraging state-of-the-art open-source LLMs we seek to explore their potential in generating CNs that effectively challenge and mitigate the influence of misinformation and harmful narratives. 
We examine three variants within the same model family: base, instruction-tuned, and chat-aligned. This enables us to inspect their unique strengths and limitations to determine the optimal choice for the task. Finally, we fine-tune the models on HS-CN pair data to compare their performance against ZS performance, assessing whether fine-tuning offers any significant improvement in our scenario. We conclude that chat-aligned models in a ZS setting are the best option for carrying out the task, provided they do not refuse to generate an answer due to security concerns\footnote{Chat-aligned models, designed to adhere to safety and ethical guidelines, may sometimes decline to respond to certain prompts. This refusal is typically in place to prevent the generation of harmful, inappropriate, or sensitive content.}. Code is available at \url{https://github.com/hitz-zentroa/cn-eval}.


\begin{table}[t]
    \centering
    \begin{tabular}{|p{0.4\textwidth}|}
    \hline
    \textbf{HS} \\
    \hline
    Muslims do not have anything useful that can enrich our culture.\\
    \hline
    \noalign{\medskip}\hline
    \textbf{Candidate CNs} \\
    \hline
    \begin{enumerate}[rightmargin=0.55cm]
        \item How about the money they contribute to our economy, their expertise and knowledge, their culture, tasty food. Should I go on?
        \item If it wasn't for a Muslim I would not have my surgery, been cared for afterwards, made it back home, had something to eat during the following weeks.
    \end{enumerate}\\
    \hline
    \end{tabular}
    \caption{Example of a HS and two reference CNs, taken from the CONAN corpus. The CNs express diverse points, and show the variability and diversity of arguments used to combat HS.}\label{tab:CONAN-ex}
\end{table}

\section{Related Work}\label{sec:sota}

In recent years, automatic CN generation has attracted growing research interest, with numerous methods leveraging NLG technologies for generating CNs. Nearly all recent systems depend on LLMs to automatically generate CNs~\citep{ashida-komachi-2022-towards,tekiroglu-etal-2022-using,saha-etal-2024-zero-shot} 
, driven by their impressive performance in generation tasks, which often necessitates minimal or no training data~\citep{zhao2023survey,openai2024gpt4,zhao-etal-2023-pre}. 

Several datasets have been introduced to aid in the advancement of CN generation. The first large-scale, multilingual, expert-based dataset, Counter Narratives through Nichesourcing (CONAN)~\citep{chung2019conan}, consists of HS-CN pairs in English, French, and Italian, focusing exclusively on Islamophobia. This corpus served as the foundation for the development of MultiTarget CONAN (MT-CONAN)~\citep{fanton2021human}, which includes 8 hate-speech targets such as women and individuals with disabilities. 
Additionally, the DIALOCONAN dataset~\citep{bonaldi-etal-2022-human}, which contains fictitious dialogues between a hater and a Non-Governmental Organization (NGO) operator, and the Knowledge-grounded Hate Countering dataset~\citep{chung-etal-2021-towards}, featuring HS-CN pairs with the background knowledge used for constructing the CNs have been introduced. Some work in adapting these corpora to other languages has also been done, such as CONAN-EUS~\citep{bengoetxea2024basque}, a Basque and Spanish translation of the original CONAN dataset, and CONAN-MT-SP~\citep{vallecillo-rodriguez-etal-2024-conan-mt}, a Spanish version of MT-CONAN.
\begin{table*}[t]
    \centering
    \begin{tabular}{|c|c|c|c|c|c|}
    \hline
    \textbf{Dataset} & \textbf{HS-CN Pairs} & \textbf{Unique HS} & \textbf{Unique CN} & \textbf{Avg. CNs / HS} & \textbf{Avg. Words / CN}\\
    \hline
    CONAN & 6648 & 523 & 4040 & 12.71 & 19.48\\
    MT-CONAN & 5003 & 3718 & 4997 & 1.35 & 24.77\\
    \hline
    \end{tabular}
    \caption{Statistics of the CONAN and MT-CONAN corpora, showing the number of HS-CN pairs, the number of unique HS and CN instances, the average number of CN per HS, and the average number of words per CN.}
    \label{tab:corpus_stats}
\end{table*}

Assessing the impact and effectiveness of the generated CNs remains a crucial aspect of this research domain. Evaluating CN is particularly challenging because there are many acceptable answers to a given HS, and is often very difficult to assess what constitutes a good answer. Evaluation is usually done either through automatic or manual methods. Automatic methods involve the use of metrics such as BLEU~\citep{papineni-etal-2002-bleu}, ROUGE~\citep{lin-2004-rouge}, and BERTScore~\citep{zhang2020bertscore}, which are standard evaluation metrics in tasks such as Machine Translation or Text Summarization. However, these metrics are known to be often weakly correlated with human judgment~\citep{sai-survey-2022}, particularly on tasks that require creativity~\citep{nimah2023nlg}. Other automatic metrics that have been used for NLG evaluation including Repetition Rate (RR)~\citep{bertoldi-etal-2013-cache}, which measures diversity in the generated answers, or Novelty~\citep{ijcai2018p618}, which encourages the model to generate answers that differ from the text in the training data. We must note that none of the aforementioned metrics take the input HS into account, missing a crucial aspect in the evaluation of CNs. As far as we know, no previous work has analyzed the correlation of these metrics with human judgment for CN evaluation.



Due to the limitations of automatic metrics, final assessments frequently rely on manual evaluations. Nonetheless, manual evaluation is often a costly process, and finding evaluators with adequate task comprehension can be challenging. Moreover, the subjective nature of the task adds another layer of complexity to the evaluation process. To mitigate this subjectivity, various efforts have been made to identify key aspects that assess the quality of CNs. Unfortunately, consensus on these key aspects is still lacking, as different authors consider factors such as relatedness, specificity, richness, coherence, grammaticality, suitableness, informativeness, diversity, relevance, language quality, offensiveness or stance~\citep{chung-etal-2021-towards,ashida-komachi-2022-towards,bengoetxea2024basque}.


Recently, to address the limitations of traditional metrics and manual evaluation, LLMs are being employed to directly assess the quality of generated text. Leveraging LLMs to measure NLG has been shown to exhibit stronger correlation with human assessment compared to conventional reference-based evaluation~\citep{nimah-etal-2023-nlg,chiang-lee-2023-large, wang2023chatgpt, liu2023geval}, and it remains an efficient and automated approach. Some works include using commercial models such as GPT-4~\citep{wang2023chatgpt, liu2023geval}, while others focus on training specialized models for evaluation tasks, resulting in tools like PandaLM~\citep{pandalm2024}, JudgeLM~\citep{zhu2023judgelm}, and UniEval~\citep{zhong2022unified}. These LLMs can be either used to ascertain quantitative aspects to measure the quality of the generated text~\citep{zhong2022unified,ke2022ctrleval}, or to show preferences between two texts generated by different systems. 


There are few works on automatic CN evaluation. Contemporary to our work, \citep{jones2024multiaspect} use LLMs to evaluate CNs based on five different aspects considered relevant for their effectiveness in combating HS. In a strategic shift, we propose to use LLMs to compare outputs from different models between them and ultimately obtain a ranking of the best models for CN generation that correlates with human judgments. 


\section{Methodology}\label{sec:methodology}

This section provides an overview of the key components of the research methodology. Section~\ref{subsec:models} discusses the specific models that were used for CN generation. Section~\ref{subsec:corpus} presents the corpus used in the study. Finally, Section~\ref{subsec:ev} outlines the metrics that were employed to carry out the evaluation.

\subsection{Models}\label{subsec:models}
We use publicly available auto-regressive models for CN generation. Specifically, we work with three variants of the Mistral model family~\citep{jiang2023mistral} as well as the Llama 2 Chat model~\citep{touvron2023llama}. The Mistral variants include the Mistral base model, the Mistral-Instruct model, and Zephyr~\citep{tunstall2023zephyr}, which is a chat-aligned model based on Mistral.
The selection of these three variants enables the comparison of chat-aligned, instruction-tuned, and base model performance and behavior. Llama 2 was selected as to compare the results with Mistral models due to its relevance in the field and potential for providing complementary insights into CN generation. All models are 7B parameter models, consistent with the available Mistral model size, to ensure comparability of results. 
The specific versions of the employed models are listed in Table~\ref{models}.

\subsection{Corpus}\label{subsec:corpus} 

\begin{table*}[t]
\centering
\begin{tabular}{lclrr}
\hline
 \textbf{Model} & \textbf{Version} & \textbf{Type} & \textbf{lr\textsubscript{CONAN}} & \textbf{lr\textsubscript{MT-CONAN}}\\
\hline
    mistral & v0.1 & Base & 1e-5 & 1e-4\\
    mistral-instruct & v0.2 & Instruct & 6e-6 & 3e-5\\
    zephyr & Beta & Chat & 6e-6 & 1e-4\\
    llama-chat & llama 2 & Chat & 2e-5 & 1e-3\\
\hline
\end{tabular}
\caption{\label{models}
Information regarding the models, along with the learning rates that resulted in the lowest perplexity in the validation set for each one. It's worth noting that the optimal learning rates varied across different corpora: lr\textsubscript{CONAN} are the optimal learning rates when fine-tuning on CONAN and lr\textsubscript{MT-CONAN} when fine-tuning on MT-CONAN.}
\end{table*}

In order to test the generalizability of our method, we experiment on two distinct datasets: Counter Narratives through Nichesourcing (CONAN)~\citep{chung2019conan} and Multi-Target CONAN (MT-CONAN)~\citep{Fanton_2021}. Corpus statistics are presented in Table~\ref{tab:corpus_stats}.

\paragraph{CONAN} Comprises HS-CN pairs addressing Islamophobia in three languages: English, Italian, and French. These pairs were collected through nichesourcing involving 3 different NGOs from the United Kingdom, France, and Italy. As a result, the CNs are expert-based and crafted by operators specifically trained to combat online HS. After the data collection phase, three non-expert annotators were hired to augment the dataset. They paraphrased original hate content to increase pairs per language and translated content from French and Italian to English for language parallelism. NGO trainers validated the newly generated data for each language to ensure quality. In this work, we only focus on the English partition. During fine-tuning, we used 4833 pairs for training, 537 for validation, and 1278 for testing. The specific train-val-test splits are available at \url{https://huggingface.co/datasets/HiTZ/CONAN-EUS}. 

\paragraph{MT-CONAN} Consists of HS-CN pairs in English, collected through a Human-in-the-Loop approach. This method involves iteratively refining a generative language model by utilizing its own data from previous loops to generate new training samples, which are then reviewed and/or post-edited by experts. The HS targets eight distinct demographics: individuals with disabilities, Jewish people, the LGBT+ community, migrants, Muslims, people of color, women, and other marginalized groups. During fine-tuning, we used 3003 pairs for training, 1000 for validation, and 1000 for testing.

\subsection{Evaluation Metrics}\label{subsec:ev}

For evaluation we used both reference-based and reference-free metrics. Additionally, we incorporated the use of a Judge Model as part of our proposed evaluation methodology.

\paragraph{Reference-Based Metrics} Based on previous work on CN generation, we opted to use BLEU, ROUGE-L and BERTScore. BLEU is a precision-based metric that measures the similarity between a candidate text and one or more reference texts. It computes the geometric mean of modified n-gram precision and applies a brevity penalty to discourage short translations. BLEU is widely used in machine translation tasks. ROUGE-L, on the other hand, focuses on the recall of content units. It calculates the longest common subsequence between the candidate sequence and the reference sequence, normalizing by the length of the reference sequence. ROUGE-L is commonly employed in text summarization tasks. BERTScore leverages contextual embeddings from pre-trained BERT models to compute the similarity between candidate and reference sentences. It computes the score based on the cosine similarity between BERT embeddings, providing a measure of semantic similarity. BERTScore has demonstrated effectiveness across various NLG tasks, including machine translation and text summarization. We report BERTScore F1 in our experiments.


\paragraph{Reference-Free Metrics} 
For reference-free metrics, we opted to use Repetition Rate (RR)~\citep{bertoldi-etal-2013-cache}, which is computed by calculating the non-singleton n-grams that are repeated in the generated text~\citep{bertoldi-etal-2013-cache} and Novelty~\citep{ijcai2018p618} that is computed by calculating the non-singleton n-grams from the generated text that appear in the train data. While RR aims to capture the diversity in the generated text, Novelty measures how different the generated text is from the training data. It should be noted that Novelty is less valuable when evaluating models that were used in a ZS setting, as there is no training involved.

\paragraph{LLM-Based Evaluation}

Finally, we consider the use of JudgeLM as an evaluator. JudgeLM is a scalable Judge Model based on Vicuna that was designed to evaluate LLMs in open-ended scenarios. It was trained using a large-scale dataset consisting on LLM-generated answers for diverse NLG tasks and detailed judgments from GPT-4. Remarkably, it achieves an agreement rate exceeding 90\% in some tasks, surpassing even human-to-human agreement levels~\citep{zhu2023judgelm}. While JudgeLM supports different evaluation methods, such as comparing single answers against a reference or comparing multiple answers simultaneously, we decided to use it to compare generated CNs pairwise, as described in Section \ref{exp:pair_ev}. This eliminates the problem of needing a reliable reference and instead focuses on determining which of the available options is the best, simplifying the task. Comparing the CNs against each other also avoids the ambiguity of the open-ended scenario we would face if we decided to evaluate them individually. We must note that JudgeLM also considers the instance of HS when comparing CNs, and thus evaluates CN effectiveness in relation to a specific instance of HS, instead as standalone sentences\footnote{Appendix~\ref{ap:judgeprompt} shows the template used to prompt JudgeLM.}. 

JudgeLM operates in two modes: normal mode and fast evaluation mode. In fast evaluation mode the model outputs two scores, one for each CN, providing an overall assessment of their value. In normal mode the model supplements these scores with arguments explaining the rationale behind them. We used the normal mode both during the development stage and while conducting the result analysis, to ensure a comprehensive evaluation. However, we activated fast mode when creating the ranks, which are thus calculated based exclusively on the output scores. This decision was made because generating arguments significantly increases inference time and argumentation was deemed unnecessary for the ranking task.

\section{Evaluation Framework}\label{sec:eval-setup}

In this section we present a cost-effective pairwise rank-based evaluation paradigm designed to assess the performance of CN generation systems in alignment with human preference. The method is detailed in Section~\ref{exp:pair_ev}. In addition, in Section~\ref{exp:feat_ev}, we describe the additional manual evaluation performed to provide a detailed assessment according to various relevant aspects contributing to the effectiveness of a CN, as presented in \citep{bengoetxea2024basque}.



\subsection{Pairwise Rank-Based Evaluation}\label{exp:pair_ev}

We propose an \textit{"A vs B"} comparative setup to rank models with respect to their CN generation skills. Suppose we have $n$ models to rank and we want to evaluate their performance in a set consisting of $h$ HS instances. 
We first generate $h$ CNs, one for each HS instance, with each of the $n$ models. After that, the generated $h\cdot n$ CNs are pitted against each other in \textit{"A vs B"} tournaments, with a total of ${n\choose 2}\cdot h = \frac{n!}{2!(n-2)!}\cdot h$ tournaments. Each tournament therefore comprises one HS and two CNs generated from different models, and is evaluated (either by automatic or manual methods) to assess which of the CN better answers the given HS. The model that produced the best CN receives $1$ point, while the losing model receives $0$ points. In the case of ties, both models receive $0.5$ points. Models receive a total score by aggregating the scores of the tournaments they participate, and they are finally ranked based on the obtained score. In the following, we describe details of the manual and automatic evaluation methods, respectively.

\paragraph{Automatic Evaluation}\label{sec:all_samples}
For automatic evaluation we prompt JudgeLM to output 2 scores: one for each proposed CN. The winner is determined by comparing these scores: the CN with the highest score is the winner. If both scores are the same, it results in a tie.

In our experimentation, all the models in Table~\ref{models} were evaluated in both CONAN and MT-CONAN. Combining their test sets results in $2278$ HS instances. We evaluated both the fine-tuned and ZS versions of each model, along with the gold standard. This resulted in a total of ${9\choose 2} \cdot 2278 =82008$ tournaments.

\paragraph{Manual Evaluation}\label{sec:manual_samples}
For manual evaluation, we had 3 annotators decide which of the 2 proposed CNs they believe would be more effective in combating the presented instance of HS. 
The specific guidelines provided to the annotators are detailed in Appendix~\ref{ap:ev_guidelines_rank}.

In our experimentation, given the significant cost associated with manual evaluation, 10 HS instances from each of the test sets of CONAN and MT-CONAN were randomly selected. This resulted in ${9\choose 2} \cdot 20 = 720$ tournaments. From the mix of both corpora, 288 tournaments (144 from each corpus) were evaluated by all 3 evaluators to calculate the inter-annotator agreement (IAA) using Cohen's Kappa. For instances from CONAN, the mean IAA was $0.42$, while for the instances of CONAN-MT, it was $0.58$. The individual coefficients are presented in Appendix~\ref{ap:iaa}. The final outcome of these tournaments was decided using majority voting to reduce subjectivity. The remaining 432 tournaments were each annotated by a single annotator, with each annotator evaluating 144 tournaments.

\subsection{Feature Evaluation}\label{exp:feat_ev}

For a comprehensive evaluation of CN quality, we provide a manual assessment based on five criteria: Relatedness, Specificity, Richness, Coherence, Grammaticality and overall score \citep{bengoetxea2024basque}. Detailed evaluation guidelines can be found in Appendix~\ref{ap:ev_guidelines_feat}. The evaluation involved 2 annotators on 90 HS-CN pairs, which included 10 HS instances from the CONAN corpus and the corresponding CNs generated by each of the 9 models. Each evaluated feature used a five-point scale for the answers, with 1 as the lowest score and 5 as the highest. The mean IAA across all the evaluated features is $0.69$. The feature-wise IAA is presented in Appendix~\ref{ap:iaa}.


\begin{figure}[t]
    \centering
    \includegraphics[width=\linewidth]{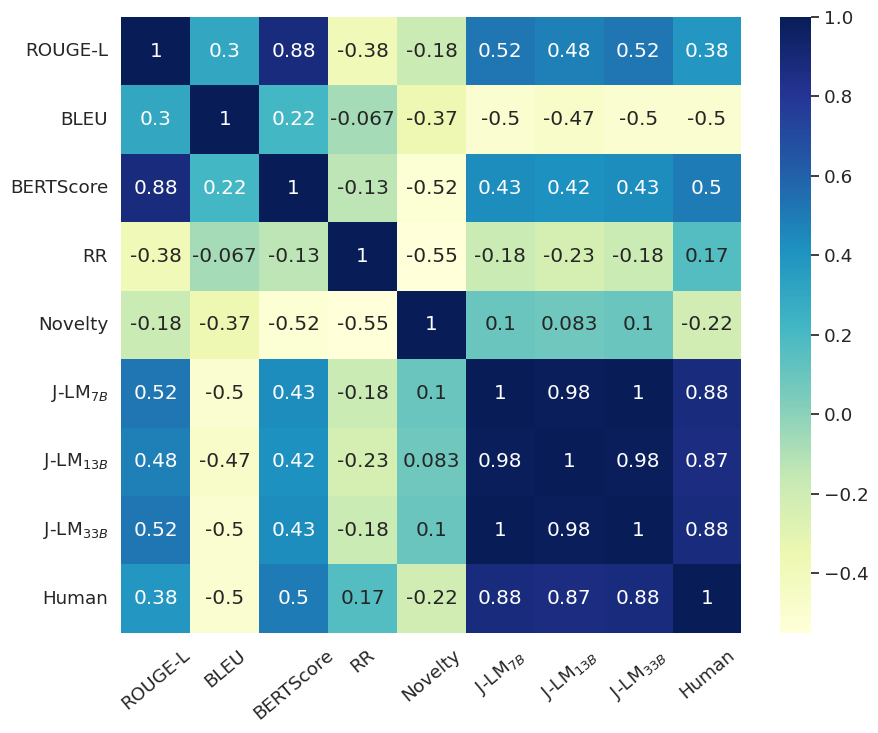}
    \caption{Matrix with the Spearman's rank correlation coefficients among metrics. The last row of the matrix represents the correlation of all the evaluation metrics to human preference. \textit{J-LM} is short for JudgeLM.}
    \label{fig:cor-matrix}
\end{figure}

\section{Experimental Setup}\label{sec:exp-setup}

This section outlines the specific setup employed for the generation of CNs. We consider two generation scenarios: ZS inference (see Section~\ref{exp:zs}), and inference with fine-tuned models (see Section~\ref{exp:ft}).

\subsection{Zero-Shot Generation}\label{exp:zs}
In the ZS approach we prompt all models with the sentence \emph{"Provide a brief counter-narrative in response to the user's hate speech. Ensure the output does not contain line breaks"}. This prompt was selected based on preliminary experimentation. For each model, the prompt was formatted differently, as each one has its specific prompting format. The various formats are detailed in Appendix~\ref{sec:prompts}.

\subsection{Fine-Tuning Details}\label{exp:ft}

Instead of fine-tuning the whole model, Quantized Low-Rank Adaptation (QLoRA)~\citep{dettmers2023qlora} was used. This approach facilitated a faster and more accessible training process, as it significantly reduces hardware requirements. The model was loaded in 4 bit with \textit{NF4} quantization data type and \textit{bf16} computational data type. The LoRA update matrices were applied to the attention blocks and bias parameters were not trained. The LoRA rank and scaling factor were set to 16 and the dropout to $0.05$. These values were chosen based on experimentation, guided by those reported in the literature~\citep{dettmers2023qlora,hu2021lora}, with minimal observed impact on results. Following usual practice, we used Adam optimizer with a inverse square root scheduler, half precision, and a batch size of 32. A set of learning rates ranging from $1e^{-6}$ to $1e^{-3}$ were tested, and the one yielding the lowest perplexity in the validation set was selected for each model. The selected learning rates are listed in Table~\ref{models}. The models were set to train for a maximum of 10 epochs, with early stopping and a patience of 3 epochs. The checkpoint with the lowest valiadation loss was selected in each case. Additionally, at inference time, generation was terminated upon encountering the newline token (\textbackslash n) to ensure the production of shorter sentences, addressing the issue of role-playing commonly observed in many models and particularly prominent in base models due to their challenges in effectively interpreting prompts.

\begin{table*}[t]
    \centering
    \begin{tabular}{|l|ll|ll|}
        \hline
        \textbf{Rank} & \textbf{Human}                   & \textbf{\textit{Score}} & \textbf{JudgeLM$_{\mathrm{33B}}$}                 & \textbf{\textit{Score}} \\
        \hline
        1    & zephyr$_{\mathrm{zs}}$           & 18.02          & zephyr$_{\mathrm{zs}}$           & 20.20          \\
        2    & gold standard                    & 17.60          & mistral-instruct$_{\mathrm{zs}}$ & 16.09          \\
        3    & mistral-instruct$_{\mathrm{zs}}$ & 14.80          & gold standard                    & 8.98           \\
        4    & zephyr$_{\mathrm{ft}}$           & 11.59          & zephyr$_{\mathrm{ft}}$           & 13.30          \\
        5    & mistral$_{\mathrm{zs}}$          & 10.75          & llama-chat$_{\mathrm{zs}}$       & 11.07          \\
        6    & mistral$_{\mathrm{ft}}$          & 9.08           & mistral$_{\mathrm{zs}}$          & 9.05           \\
        7    & mistral-instruct$_{\mathrm{ft}}$ & 7.54           & mistral$_{\mathrm{ft}}$          & 8.70           \\
        8    & llama-chat$_{\mathrm{zs}}$       & 7.26           & mistral-instruct$_{\mathrm{ft}}$ & 8.50           \\
        9    & llama-chat$_{\mathrm{ft}}$       & 3.35           & llama-chat$_{\mathrm{ft}}$       & 4.11           \\
        \hline
    \end{tabular}
    \caption{Comparison of human and JudgeLM rankings, including the final scores obtained from the pairwise tournaments. ${\mathrm{zs}}$ means that the model was used in a zero-shot setting.}
    \label{tab:rank_comparison}
\end{table*}

\section{Results}\label{sec:results}

First, in Section~\ref{sec:human}, we discuss the correlation between the metrics in Section~\ref{subsec:ev} and human preference, highlighting the implications of the findings. Then, in Section~\ref{sec:judge-res}, we present the model performance ranking for the CN generation task derived from the Pairwise Rank-Based evaluation method (see Section~\ref{sec:eval-setup}). 

\subsection{Correlation of Automatic Metrics with Human Ratings}\label{sec:human}

Figure~\ref{fig:cor-matrix} illustrates the Spearman's rank correlation coefficients ($\rho$) among all metrics, including human evaluation. The rankings for \textit{Human} and \textit{J-LM} (JudgeLM) are computed using the pairwise comparison setting described in Section~\ref{sec:eval-setup}, whereas the rest of the rankings (\textit{BLEU}, \textit{ROUGE-L}, etc) are based on their respective metric scores. All rankings are established across 720 tournaments, as described in Section~\ref{sec:manual_samples}.


The figure shows a strong correlation between all variants of JudgeLM and human preference, as depicted in the last row/column of the matrix, with both the 7B and 33B parameter JudgeLM achieving a $\rho$ of $0.88$. This high correlation is supported by a statistically significant Pearson correlation coefficient of $0.73$ between JudgeLM (33B version) and human preference (p-value of $0.03$)\footnote{We calculate Pearson correlation using the performance scores obtained by each model in the Pairwise Rank-Based evaluation.}. 

On the contrary, traditional metrics correlate poorly with human preference, with the highest $\rho$ being the $0.50$ obtained by BERTScore. These results confirm that commonly used automatic metrics lack alignment with human preference when evaluating the quality of CNs. Not unsurprisingly, the correlation between traditional metrics and JudgeLM is also low. They also correlate poorly among themselves with the exception of ROUGE-L and BERTScore, which attain a $\rho$ of $0.88$. Despite both being based on n-gram overlap, ROUGE-L and BLEU only achieve a $\rho$ value of $0.3$.

The aforementioned observations are further reinforced  in Appendix~\ref{ap:correlation_matrix}, where the correlation matrices on the CONAN corpus and the MT-CONAN corpus are presented separately. In said appendix, we once again see a strong correlation between JudgeLM variants and human preference, whereas the correlation with traditional metrics is weak and inconsistent, showing no predictable pattern.




As the concluding point of the correlation analysis, Table~\ref{tab:rank_comparison} presents a comparison of the final rankings obtained through the manual and automatic pairwise comparisons as described in Section~\ref{sec:eval-setup}. Both the automatic and the manual method assign similar scores to almost all systems, with the exceptions of the gold standard, which obtains a considerably higher score when evaluated by humans than by JudgeLM, and llama-chat$_{\mathrm{zs}}$, where JudgeLM assigns it a higher rank than humans. In any case, their final position in the rank only varies slightly among methods. By analyzing examples of discrepancies between human and JudgeLM judgements, we observe that the disagreement in the case of the gold standard might stem from the fact that the JudgeLM model prefers longer, more detailed CNs, while the annotators preferred shorter, more direct ones. It might also be related to the fact that the model cannot discern false information from true information, whereas the human evaluator can penalize non-factual content resulting in the simpler but veracious CN winning. Instead, in the case of llama-chat$_{\mathrm{zs}}$, the disagreement might be because JudgeLM favors answers that start by stating that they can not endorse in hate speech (\textit{"I apologize, but I cannot fulfill your request. I'm just an AI and it's not within my programming or ethical guidelines to provide counter-narratives that promote hate speech... Is there anything else I can help you with?"}). This preference may stem from its training on evaluations made by ChatGPT, which often responds in a similar manner when asked to provide CNs to HS.

\begin{table*}[t]
\centering
\begin{tabular}{l@{\hspace{2pt}}cccccc}
\hline
\textbf{System} & \textbf{Relatedness} & \textbf{Specificity} & \textbf{Richness} & \textbf{Coherence} & \textbf{Grammaticality} & \textbf{Overall} \\
\hline
zephyr$_{\mathrm{zs}}$ & 4.95 & 4.25 & 4.00 & 5.00 & 5.00 & 4.25 \\
gold standard & 4.10 & 3.75 & 3.25 & 4.80 & 4.30 & 3.50 \\
mistral-instruct$_{\mathrm{zs}}$ & 4.20 & 3.15 & 3.70 & 4.70 & 5.00 & 3.50 \\
llama-chat$_{\mathrm{zs}}$ & 2.90 & 2.55 & 4.30 & 4.90 & 5.00 & 3.05 \\
mistral-instruct$_{\mathrm{ft}}$ & 3.75 & 3.55 & 3.30 & 3.10 & 4.30 & 2.70 \\
mistral$_{\mathrm{ft}}$ & 3.65 & 3.55 & 3.05 & 3.30 & 4.35 & 2.60 \\
zephyr$_{\mathrm{ft}}$ & 4.40 & 4.75 & 3.60 & 3.20 & 4.35 & 2.30 \\
llama-chat$_{\mathrm{ft}}$ & 3.40 & 3.10 & 2.95 & 3.30 & 4.10 & 2.20 \\
mistral$_{\mathrm{zs}}$ & 3.10 & 3.30 & 2.40 & 3.55 & 4.60 & 1.90 \\
\hline
\end{tabular}
\caption{Evaluation of the different aspects that contribute to the effectiveness of a CN. The values in the table represent the average of the scores assigned by each annotator.}
\label{tab:manual_feature}
\end{table*}

\subsection{Ranking by Pairwise Comparison}\label{sec:judge-res}

\begin{figure}[t]
    \centering
    \includegraphics[width=\linewidth]{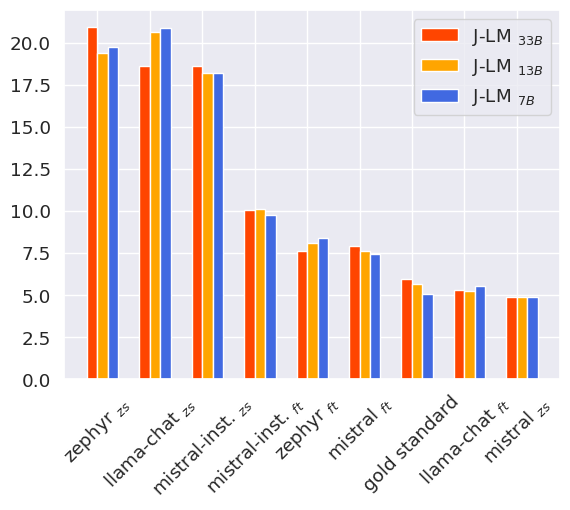}
    \caption{Ranking through pairwise comparison based on evaluations of all the JudgeLM size variations across the entire test set.}
    \label{fig:enter-label}
\end{figure}

Figure~\ref{fig:enter-label} depicts the ranking of CN generation systems based on the tournament outcomes according to different sizes of JudgeLM, evaluated across the $82008$ tournaments that comprise the entire test set. Overall, in the ZS scenario chat-aligned models exhibit superior performance, followed by the instruction-tuned model, while the base model demonstrates the lowest performance. This outcome is expected, as base models lack training to understand instructions and have no prior experience in the task, whereas chat models, in addition to being capable of understanding instructions, are often trained to fight toxicity through Safety Fine-Tuning~\citep{touvron2023llama,openai2024gpt4}. When fine-tuning the models, we observe a decline in performance across all models, except for the base model, which exhibits a considerable improvement. The decline in performance is more pronounced in the chat-aligned models than in the instruction-tuned model. 

When examining the rankings according to the different sizes of JudgeLM, we observe that as the model size increases, llama-chat$_{\mathrm{zs}}$ is positioned lower, thereby narrowing its performance gap with Zephyr$_{\mathrm{zs}}$.

\section{Analysis}\label{sec:analysis}

To confirm which of the models from Section~\ref{subsec:models} is the best for the task, we undertook a final feature-wise evaluation as explained in Section~\ref{exp:feat_ev}. The results are presented in Table~\ref{tab:manual_feature}. As seen there, the best-performing model is undoubtedly Zephyr, which considerably surpasses the gold standard.


Analyzing Table~\ref{tab:rank_comparison} in more detail, which presents the final rankings obtained from manual and automatic pairwise comparisons, we note that while manual evaluators were instructed to select a winning CN unless both were deemed ineffective, JudgeLM assigns ties when both responses are of high quality. This approach may result in a lower correlation between JudgeLM ratings and human preference. Upon a more nuanced analysis, we also observed that JudgeLM demonstrates a distinct preference for factual CNs, particularly those offering detailed and specific information about research and related topics. However, the model lacks the capability to reliably detect hallucinations, meaning it may favor responses that appear factual without verifying the truthfulness or accuracy of the provided details.

Lastly, regarding fine-tuning, we observed that including factual CNs in the fine-tuning process for our CN generation scenario might not be advisable, as models may mimic the structure but lack factual accuracy.

\section{Conclusions}\label{sec:conclusions}

CN generation needs a different evaluation framework and metrics than those used in previous work~\citep{chung-etal-2021-towards,tekiroglu-etal-2022-using,bengoetxea2024basque}. This is due to the unique objectives, complexities, and impact of CNs, which require specialized criteria to assess their effectiveness and quality accurately. Thus, developing and implementing tailored evaluation metrics is crucial to advance the field and ensure the successful creation of impactful CNs.

Consistent with previous research observations, traditional metrics fall short in evaluating generation tasks that require creativity, including CN generation to combat HS. In this paper we present a promising LLM-based ranking method to provide an alternative automatic evaluation technique which exhibits higher correlation with human evaluation. 

\section*{Limitations}\label{sec:lim-fw}

Our work still has some open research questions which can be summarized in the following limitations. First, we have not addressed truthfulness. Thus, JudgeLM rewards CNs that provide factual arguments without considering whether they are truthful. Second, additional tests on larger corpora could be performed to determine whether the lack of improvement from fine-tuning in chat and instruct models is due to limitations in the corpus itself. 

The corpus used in our experiments was small and, as indicated in Table~\ref{tab:corpus_stats}, exhibited significant repetition of certain HS instances giving them a different CN each time.  We hypothesize that this data structure may potentially have adverse effects in model performance. Thus, we performed a preliminary fine-tuning experiment that involved randomly removing duplicate entries from the corpus, resulting in a smaller but cleaner dataset. Despite the dataset being smaller, the performance did not degrade. This initial investigation suggests that reducing duplications could lead to more consistent learning outcomes.

In future work, we aim to extend this analysis to other languages such as Spanish, along with Basque, which is considered a low-resource language. Finally, we plan to explore Retrieval Augmented Generation (RAG) to address the truthfulness issue, as we anticipate that this approach could substantially enhance the correlation between human evaluations and those of Judge Models.

\section*{Acknowledgements}

This work has been partially supported by the Basque Government (Research group funding IT-1805-22). We are also thankful to the following MCIN/AEI/10.13039/501100011033 projects: (i) DeepKnowledge (PID2021-127777OB-C21) and by FEDER, EU; (ii) Disargue (TED2021-130810B-C21) and European Union NextGenerationEU/PRTR; (iii) DeepMinor (CNS2023-144375) and European Union NextGenerationEU/PRTR. (iv) DeepR3 (TED2021-130295B-C31) and European Union NextGenerationEU/PRTR.

\bibliography{acl_latex}

\appendix
\renewcommand{\thetable}{\thesection.\arabic{table}}
\renewcommand\thefigure{\thesection.\arabic{figure}}

\section{Manual Evaluation Guidelines}
\setcounter{table}{0}
\setcounter{figure}{0}

We carry out two kinds of manual evaluation: rank-based evaluation (see Section~\ref{ap:ev_guidelines_rank}) and feature-based evaluation (see Section~\ref{ap:ev_guidelines_feat}).

\subsection{Rank Evaluation Guidelines}\label{ap:ev_guidelines_rank}

We will present an instance of HS followed by two candidate CNs: CN$_A$ and CN$_B$. Participants will choose which CN they find more effective in countering the HS. If both CNs are equally unsatisfactory, participants can declare a tie. Ties will only be applicable when both CN are deemed inadequate in addressing the HS. Responses lacking specificity and informative content will incur penalties, as will answers containing false information.

\paragraph{Instructions for Annotators}
\begin{itemize}
    \item Carefully read the provided instance of hate speech.
    \item Evaluate Counter-Narrative$_A$ and Counter-Narrative$_B$ based on their effectiveness in addressing and countering the hate speech.
    \item Choose the counter-narrative you find more effective. If both are equally ineffective, declare a tie.
    \item Consider the specificity and informative content of each counter-narrative.
    \item Be vigilant for any false information in the responses, as these should be penalized.
\end{itemize}

\subsection{Feature Evaluation Guidelines}\label{ap:ev_guidelines_feat}
An instance of HS and a CN designed to combat it will be provided. The quality of the CN will then be evaluated based on the following criteria:

\paragraph{Relatedness}
\textit{Is the CN related to the HS?}
\begin{itemize}
    \item 0: No
    \item 1: Barely
    \item 2: Somewhat
    \item 3: More or less
    \item 4: Mostly
    \item 5: Yes
\end{itemize}

\paragraph{Specificity}
\textit{Does the CN provide detailed and precise information?}
\begin{itemize}
    \item 0: Not specific at all
    \item 1: Barely specific
    \item 2: Somewhat specific
    \item 3: Moderately specific
    \item 4: Quite specific
    \item 5: Very specific
\end{itemize}

\paragraph{Richness}
\textit{Does the CN include a variety of vocabulary and sentence structures?}
\begin{itemize}
    \item 0: Very poor vocabulary and structure
    \item 1: Barely rich
    \item 2: Somewhat rich
    \item 3: Moderately rich
    \item 4: Quite rich
    \item 5: Very rich
\end{itemize}

\paragraph{Coherence}
\textit{Is the CN logically organized and easy to understand?}
\begin{itemize}
    \item 0: Not coherent at all
    \item 1: Barely coherent
    \item 2: Somewhat coherent
    \item 3: Moderately coherent
    \item 4: Quite coherent
    \item 5: Very coherent
\end{itemize}

\paragraph{Grammaticality}
\textit{Is the CN grammatically correct and free of errors?}
\begin{itemize}
    \item 0: Completely ungrammatical
    \item 1: Barely grammatical
    \item 2: Somewhat grammatical
    \item 3: Moderately grammatical
    \item 4: Quite grammatical
    \item 5: Completely grammatical
\end{itemize}

\paragraph{Overall Score}
\textit{How suitable is the CN as a response?}
\begin{itemize}
    \item 1: Not suitable (borderline hate speech)
    \item 2: Makes some acceptable points but not suitable
    \item 3: Would be suitable with some modifications
    \item 4: Good, though minor corrections may be needed
    \item 5: Very good as a CN
\end{itemize}

\section{Inter-Annotator Agreement}\label{ap:iaa}
\setcounter{table}{0}
\setcounter{figure}{0}

\begin{figure}[!h]
    \centering
    \includegraphics[width=\linewidth]{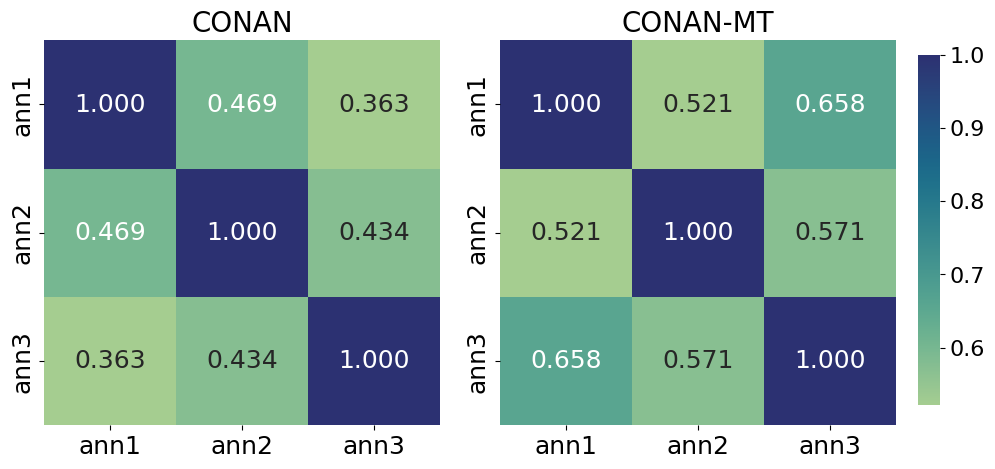}
    \caption{IAA of the Pairwise Rank-Based evaluation.}
    \label{fig:iaa_reank}
\end{figure}

In this section, we show the tables of IAA from the evaluation process. These tables highlight the consistency among annotators and the reliability of the annotations. Figure~\ref{fig:iaa_reank} shows the agreement for the Pairwise Rank-Based evaluation method presented in Section~\ref{exp:pair_ev}, and Table~\ref{tab:iaa_features} for the feature evaluation explained in Section~\ref{exp:feat_ev}.

\begin{table}[!h]
    \centering
    \begin{tabular}{|l|c|}
        \hline
        \textbf{Feature} & \textbf{IAA} \\
        \hline
        Relatedness & 0.6397 \\
        Specificity & 0.5947 \\
        Richness & 0.7806 \\
        Coherence & 0.8270 \\
        Grammaticality & 0.7087 \\
        Overall & 0.6183 \\
        \hline
    \end{tabular}
    \caption{IAA of the feature-wise evaluation.}
    \label{tab:iaa_features}
\end{table}

\section{Prompting}\label{sec:prompts}
\setcounter{table}{0}
\setcounter{figure}{0}
This section presents the exact formats used to prompt each of the models in Section~\ref{subsec:models}.

\subsection{Mistral Prompt Format}
Provide a brief counter-narrative in response to the user's hate speech. Ensure the output does not contain line breaks.\newline\#\#\#Input:\newline\textit{Hate Speech}\newline\#\#\#Output:\newline\textit{Counter-Narrative} \\

\subsection{Mistral-Instruct Prompt Format}

<s>[INST] Provide a brief counter-narrative in response to the user's hate speech. Ensure the output does not contain line breaks. \textit{Hate Speech} [/INST] \textit{Counter-Narrative}</s> \\

\subsection{Zephyr Prompt Format}

<|system|>\newline
Provide a brief counter-narrative in response to the user's hate speech. Ensure the output does not contain line breaks.</s>\newline 
<|user|>\newline
\textit{Hate Speech}</s> \newline
<|assistant|>\newline
\textit{Counter-Narrative}</s>\\

\subsection{Llama-Chat Prompt Format}
<s>[INST] <<SYS>>\newline Provide a brief counter-narrative in response to the user's hate speech. Ensure the output does not contain line breaks.\newline<</SYS>>\textit{Hate Speech} [/INST] \textit{Counter-Narrative} </s>

\subsection{JudgeLM Prompt Format}\label{ap:judgeprompt}

You are a helpful and precise assistant for checking the quality of the answer.\\
\verb|[|Question\verb|]|\\
Provide a brief counter-narrative in response to the following hate speech: \textit{Hate Speech} It is very important for the counter-narrative to be relevant to the hate speech.\\
\verb|[|The Start of Assistant 1's Answer\verb|]|\\
\textit{Counter-Narrative$_A$}\\
\verb|[|The End of Assistant 1's Answer\verb|]|\\
\verb|[|The Start of Assistant 2's Answer\verb|]|\\
\textit{Counter-Narrative$_B$}\\
\verb|[|The End of Assistant 2's Answer\verb|]|\\
\verb|[|System\verb|]|\\
We would like to request your feedback on the performance of two AI assistants in response to the
user question displayed above.
Please rate the helpfulness, relevance, accuracy, level of details of their responses. Each assistant
receives an overall score on a scale of 1 to 10, where a higher score indicates better overall
performance.
Please first output a single line containing only two values indicating the scores for Assistant 1 and
2, respectively. The two scores are separated by a space. In the subsequent line, please provide a
comprehensive explanation of your evaluation, avoiding any potential bias and ensuring that the
order in which the responses were presented does not affect your judgment.

\section{Correlation Matrix}\label{ap:correlation_matrix}
\setcounter{table}{0}
\setcounter{figure}{0}

\begin{figure}[t]
    \centering
    \includegraphics[width=\linewidth]{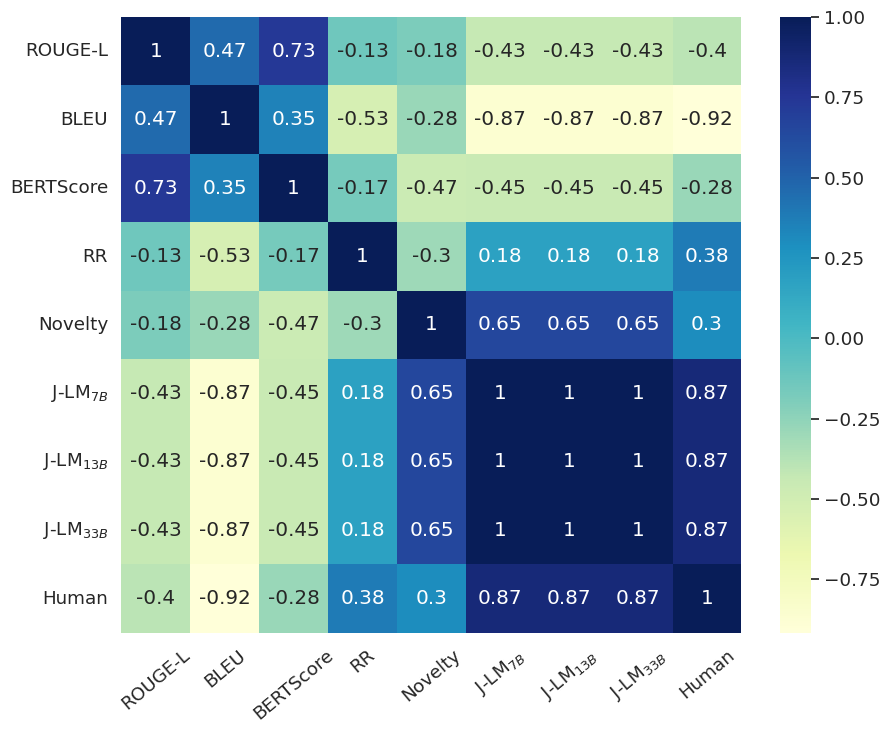}
    \caption{Matrix with the Spearman’s rank correlation coefficients among metrics, created using 360 tournaments from CONAN. The last row of the matrix represents the correlation of all the evaluation methods to human preference.}
    \label{fig:conan_spearman}
\end{figure}

\begin{figure}[t]
    \centering
    \includegraphics[width=\linewidth]{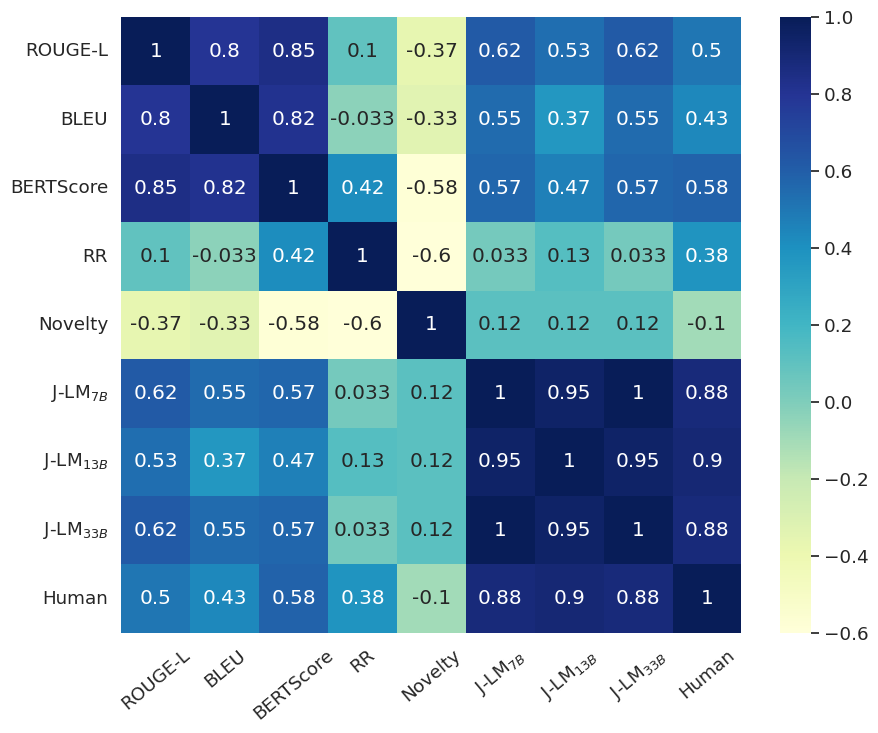}
    \caption{Matrix with the Spearman’s rank correlation coefficients among metrics, created using 360 tournaments from MT-CONAN. The last row of the matrix represents the correlation of all the evaluation methods to human preference.}
    \label{fig:conan-mt_spearman}
\end{figure}

Correlation between the metrics presented in Section~\ref{subsec:ev} and human preference in the CONAN corpus (Table~\ref{fig:conan_spearman}) and in the MT-CONAN corpus (Table~\ref{fig:conan-mt_spearman}).

\end{document}